\documentclass[pmlr,twocolumn,10pt]{jmlr} 

\usepackage{listings}
\usepackage{tcolorbox}
\tcbuselibrary{listingsutf8}
\usepackage{booktabs}
\usepackage{float} 
\usepackage{siunitx}
\newcolumntype{Y}{>{\centering\arraybackslash}X}

\usepackage[switch]{lineno}

\author{%
\Name{Sharim Khan}\Email{sharimk2@illinois.edu}\\
\addr University of Illinois, Urbana-Champaign, USA
\AND
\Name{Paul Landes}\Email{plande2@uic.edu}\\
\addr University of Illinois Chicago, Chicago, USA
\AND
\Name{Adam Cross}\Email{arcross@uic.edu}\\
\addr University of Illinois Chicago, Peoria, USA
\AND
\Name{Jimeng Sun}\Email{jimeng@illinois.edu}\\
\addr University of Illinois, Urbana-Champaign, USA
}

\theorembodyfont{\upshape}
\theoremheaderfont{\scshape}
\theorempostheader{:}
\theoremsep{\newline}

\jmlrworkshop{Machine Learning for Health (ML4H) 2025} 

\usepackage{multirow}

\usepackage{stfloats}

\title[SDoH ICD-9 Code Predictions with LLMs]{Social Determinants of Health Prediction  \\ for ICD-9 Code with Reasoning Models}

 \usepackage{soul}
\newcommand{\tl}[2][]{}

\usepackage{acro}
\DeclareAcronym{sdoh}{
  short = SDoH,
  long = Social Determinants of Health,
  short-plural-form = SDoHs
}
\DeclareAcronym{ehr}{
  short = EHR,
  long = Electronic Health Record,
}
\DeclareAcronym{llm}{
  short = LLM,
  long = large language model,
}
\DeclareAcronym{icd9}{
  short = ICD-9,
  long  = {International Classification of Diseases, Ninth Revision}
}

\newcommand{\vname}[1]{\texttt{#1}}
\newcommand{\nadms}{139}
\newcommand{\tabsize}{\small}
\newcommand{\scol}[2]{(\textbf{#1})#2}

\begin{document}

\maketitle

\begin{abstract}
\acl{sdoh}  correlate with patient outcomes but are rarely captured in structured data. Recent attention has been given to automatically extracting these markers from clinical text to supplement diagnostic systems with knowledge of patients' social circumstances.  
\Aclp{llm} demonstrate strong performance in identifying \acl{sdoh} labels from sentences. However, prediction in large admissions or longitudinal notes is challenging given long distance dependencies.
In this paper, we explore hospital admission multi-label \acl{sdoh} ICD-9 code classification on the MIMIC-III dataset using reasoning models and traditional \aclp{llm}. 
We exploit existing ICD-9 codes for prediction on admissions, which achieved a 89\% F1.
Our contributions include our findings, missing \ac{sdoh} codes in \nadms\ admissions, and code to reproduce the results.
\end{abstract}


\acresetall

\paragraph*{Data and Code Availability}  
This study used the publicly available, de-identified MIMIC-III v1.4 dataset, accessed under the PhysioNet Data Use Agreement. Access requires completion of credentialing and training as specified by PhysioNet. The full evaluation code and prompt templates are also available\footnote{\href{https://github.com/sharimk2/icd9_llm_prediction}{https://github.com/sharimk2/icd9\_llm\_prediction}}.

\paragraph*{Institutional Review Board (IRB)}  
The analyses were conducted using de-identified secondary data (MIMIC-III) and did not involve interaction with human participants. In accordance with institutional policy, this work does not constitute human subjects research and therefore did not require IRB review.  
\begin{figure}[t!]
\includegraphics[width=0.89\columnwidth,height=6.7cm]{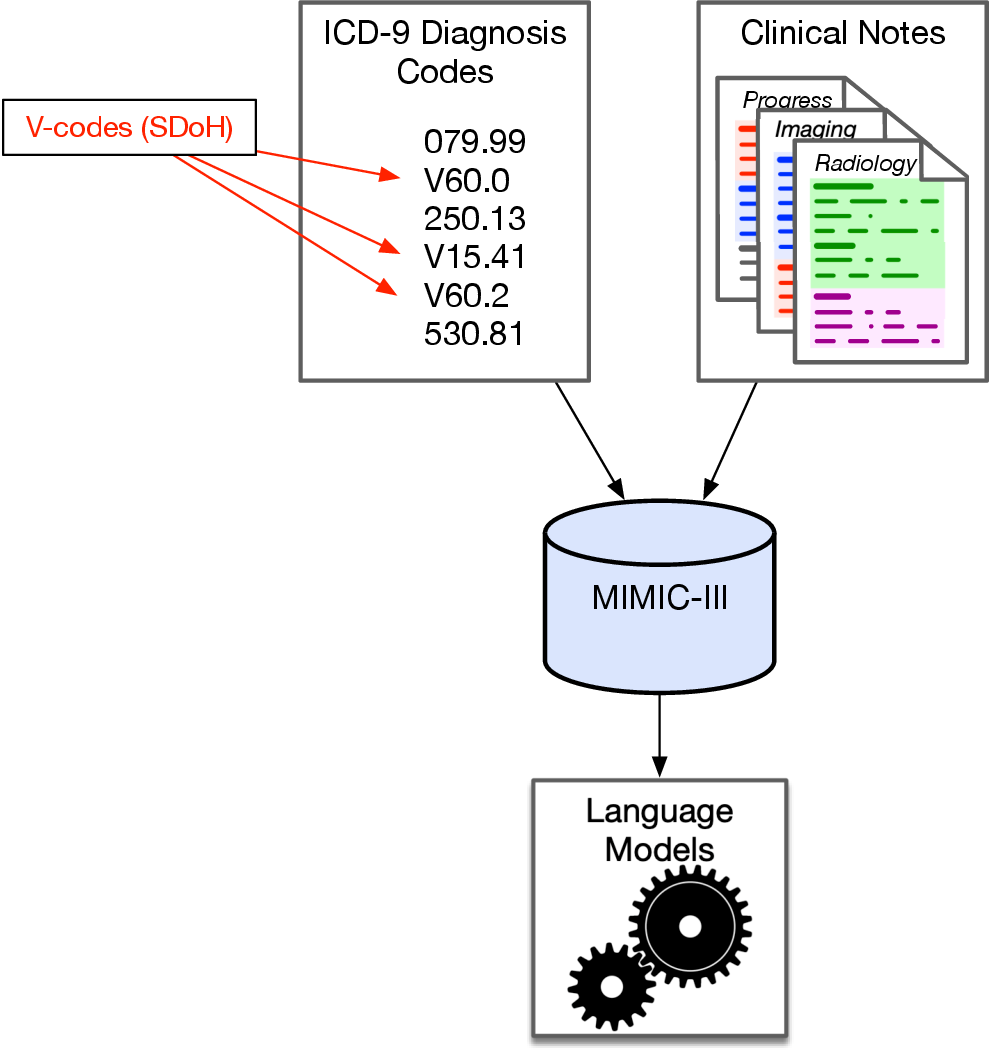}
\caption{Free text clinical notes, and ICD-9 V-codes, are used as gold labels for prediction and to identify \acs{sdoh} marked admissions.}%
\label{fig:sdoh-llm}
\end{figure}
\section{Introduction}
\label{sec:intro}

\Ac{sdoh} are non-medical factors that account for an estimated 80–90\% of modifiable contributors to health outcomes and disparities~\citep{MIInitiatives2024SDOH}. \acp{sdoh} may be adverse or protective. Adverse \acp{sdoh} are factors that create an additional social work or resource support need for patients \citep{Guevara2024LLMSDoHnpjDM}. Healthcare providers often document these factors in unstructured notes in \aclp{ehr}, which limits their utility. The \ac{icd9} classification system includes V-codes designed to capture social circumstances such as homelessness, unemployment, and inadequate family support. However, physicians diagnose \ac{sdoh} V-codes as little as 2\%, even in populations with documented social needs~\citep{Torres2017ICDSocialCodes,Hatef2019SBDHAvailability}.

This underutilization often stems from documenting social risk factors within clinical narratives, which require substantial time and expertise to diagnose as billing codes~\citep{Torres2017ICDSocialCodes}.  Reasoning-capable language models present an opportunity to predict \ac{sdoh} by using coding guidelines in prompts and training on V-codes (a subset of ICD-9 codes used by insurance companies for billing) as labels already diagnosed by physicians in the MIMIC-III dataset as shown in Figure~\ref{fig:sdoh-llm}. 

Existing \acp{llm} studies on \ac{sdoh} classify text into broad social risk categories such as housing or social support, as shown by \citet{Guevara2024LLMSDoHnpjDM}. This approach does not operate at the admission level nor predict specific billing codes. In contrast, our work directly predicts fine-grained ICD-9 V-codes from the entire set of clinical documents for an admission, bridging the gap between unstructured \ac{sdoh} documentation and structured diagnostic coding.

The semantics of the clinical notes with the diagnosed ICD-9 V-codes map to the descriptions of the labels from the annotation guide provided by \citet{Guevara2024LLMSDoHnpjDM}  as shown in Table~\ref{tab:code_mapping}. Undomiciled and unemployed are the only adverse \ac{sdoh} categories with overlap between this study and the work done by \citet{Guevara2024LLMSDoHnpjDM}.  This study evaluates the feasibility and performance of reasoning language models to predict V-codes from clinical documentation across open-source and closed frontier models.

\begin{table}[t!]
\centering
\tabsize
\begin{tabular}{lll}
\toprule
\textbf{Guevara Label} & \textbf{V-code} & \textbf{Z-code} \\
\midrule
Undomiciled & Homelessness (V60.0) & Z59\\
Unemployed & Unemployment (V62.0) & Z56 \\
\bottomrule
\end{tabular}
\caption{Mapping from \citet{Guevara2024LLMSDoHnpjDM} \ac{sdoh} labels to ICD-9 V-codes and ICD-10 Z-codes.}%
\label{tab:code_mapping}
\end{table}

\section{Related Work}
\label{sec:related}
Initial approaches to \ac{sdoh} identification employed rule-based systems and traditional machine learning methods to detect \ac{sdoh} from clinical text~\citep{Chen2020SDOHReview}. However, these methods demonstrated limited generalizability due to institution-specific documentation patterns. Subsequent advances in neural architectures, including convolutional models ~\citep{Mullenbach2018CAML} and transformer architectures such as ClinicalBERT ~\citep{clinicalbert}, improved automated medical coding performance on diagnosis and procedure codes.

Researchers have explored large language models for \ac{sdoh} mention classification in clinical notes. \citet{Guevara2024LLMSDoHnpjDM} achieved 70\% macro-F1 for adverse \ac{sdoh} classification at the sentence-level. Their work identified 93.8\% of patients with adverse \ac{sdoh} compared to 2.0\% captured by ICD-10 codes. Their taxonomy of labels maps to broad ICD-10 Z-code categories.

Hybrid architectures combining language models with traditional neural networks have demonstrated improvements in both accuracy and computational efficiency for \ac{sdoh} extraction~\citep{Landes2025HybridSDOH}.


\begin{table}[t!]
\centering
\tabsize
\begin{tabular}{lll}
\toprule
\textbf{Domain} & \textbf{Description} & \textbf{Code} \\
\midrule
\multirow{3}{*}{\shortstack[l]{Housing \& \\ Resources}}
    & Homelessness & V60.0 \\
    & Inadequate material resources & V60.2 \\
    & No family caregiver & V60.4 \\
\midrule
Employment & Unemployment & V62.0 \\
\midrule
Legal & Legal problems & V62.5 \\
\midrule
\multirow{2}{*}{Abuse History} 
    & Physical/sexual abuse & V15.41 \\
    & Emotional abuse & V15.42 \\
\midrule
Family History & Family alcoholism & V61.41 \\
\bottomrule
\end{tabular}
\caption{\Acl{sdoh} ICD-9 V-code taxonomy by domain.}
\label{tab:taxonomy}
\end{table}
\section{Dataset}


Unstructured free text clinical notes from the Medical Information Mart for Intensive Care (MIMIC-III v1.4) was used as input to several \acp{llm}.  MIMIC-III is a publicly available dataset containing de-identified health records from over 58,000 hospital admissions at Beth Israel Deaconess Medical Center between 2001 and 2012~\citep{Johnson2016MIMICIII}.  We predicted ICD-9 V-codes\footnote{MIMIC-III predates the 2015 ICD-10 transition.} identified as \ac{sdoh} by \citet{Torres2017ICDSocialCodes}, eight of which were found in the MIMIC-III dataset and listed in Table~\ref{tab:taxonomy}.


\paragraph*{Evaluation Dataset}

The evaluation dataset comprised 139 MIMIC-III admissions selected to capture varied \ac{sdoh} documentation patterns. Selection criteria included admissions with existing V-code assignments as well as admissions without \ac{sdoh} codes. We sampled 357 admissions with at least 500 characters from discharge summaries, nursing notes, and social work evaluation note categories.

\begin{table*}[t!]
\centering
\tabsize
\begin{tabular}{llcccccccccc}
\toprule
\textbf{Model} & \textbf{Parameters} & \textbf{EM} & \textbf{wP} & \textbf{wR} & \textbf{wF1} & \textbf{mP} & \textbf{mR} & \textbf{mF1} & \textbf{MP} & \textbf{MR} & \textbf{MF1} \\
\midrule
\multicolumn{12}{l}{\textit{Frontier models}} \\
GPT-3.5-turbo   & Unknown & 0.144 & 0.718 & 0.338 & 0.452 & 0.529 & 0.338 & 0.413 & 0.208 & 0.162 & 0.167 \\
GPT-4o-mini     & Unknown & 0.691 & 0.864 & \textbf{0.870} & 0.855 & 0.845 & \textbf{0.870} & 0.857 & 0.719 & \textbf{0.642} & 0.645 \\
GPT-5-mini      & Unknown & \textbf{0.722} & \textbf{0.939} & 0.847 & \textbf{0.882} & \textbf{0.939} & 0.847 & \textbf{0.891} & \textbf{0.797} & 0.608 & \textbf{0.665} \\
\midrule
\multicolumn{12}{l}{\textit{Open-source models}} \\
Llama-3.1       & 8B  & 0.103 & 0.527 & 0.765 & 0.573 & 0.412 & 0.765 & 0.536 & 0.383 & 0.521 & 0.351 \\
Llama-3.3       & 70B & 0.355 & 0.765 & \textbf{0.850} & 0.740 & 0.598 & \textbf{0.850} & 0.702 & 0.678 & \textbf{0.663} & 0.573 \\
Qwen3           & 32B & \textbf{0.655} & \textbf{0.918} & 0.770 & \textbf{0.829} & 0.925 & 0.770 & 0.840 & 0.769 & 0.577 & 0.644 \\
GPT-OSS         & 20B & 0.652 & 0.896 & 0.775 & 0.824 & \textbf{0.952} & 0.775 & \textbf{0.854} & \textbf{0.795} & 0.580 & \textbf{0.652} \\
\bottomrule
\end{tabular}
\caption{Model results metrics on \ac{sdoh} code prediction. Metrics include \scol{E}{xact} \scol{M}{atch}, \scol{P}{recision}, \scol{R}{ecall}, and F1 scores reported as \scol{w}{eighted}, \scol{m}{icro}, and \scol{M}{acro} averages. Model parameters shown in billions (B). \textbf{Bolded} measures indicate the highest performing for each subgroup.}
\label{tab:overall_models}
\end{table*}

\begin{table*}[b!]
\centering
\tabsize
\begin{tabular}{lcccc}
\toprule
\textbf{Code} & \textbf{Precision (\%)} & \textbf{Recall (\%)} & \textbf{F1 (\%)} & \textbf{Count} \\
\midrule
Homelessness (V60.0) & 100.0 & 94.8 & 97.4 & 97 \\
Inadequate material resources (V60.2) & 100.0 & 70.8 & 82.8 & 8 \\
No family caregiver (V60.4) & 95.8 & 52.8 & 67.5 & 12 \\
Unemployment (V62.0) & 95.9 & 84.5 & 89.9 & 28 \\
Legal problems (V62.5) & 56.1 & 45.2 & 50.0 & 14 \\
Physical abuse (V15.41) & 72.3 & 73.8 & 73.0 & 14 \\
Emotional abuse (V15.42) & 100.0 & 25.0 & 40.0 & 4 \\
Family alcoholism (V61.41) & 96.8 & 100.0 & 98.4 & 30 \\
\bottomrule
\end{tabular}
\caption{By code performance of GPT-5-mini on the evaluation dataset and count of admissions containing the code.}
\label{tab:per_code_performance}
\end{table*}

\section{Experimental Setup}

We identified admissions with notes containing the V-codes listed in Table~\ref{tab:taxonomy} or \acp{sdoh} not diagnosed by physicians.
We labeled these admissions with the additional V-codes for related \acp{sdoh}, which we refer to this as ``amended codes''. For example, ``He is \textit{unemployed} and
receives SSI secondary to anxiety, problems with depth
perception,'' appears in an admission only labeled with \vname{homelessness}~(V60.0).  The \textit{unemployed} token is a clear indication of a \vname{unemployment}~(V62.0), but is missing from the admission, and therefore, added as an amended code.
We evaluate the models on the amended codes to avoid penalizing for missing undiagnosed \acp{sdoh}.
The procured dataset is available via \texttt{PyHealth} ~\citep{pyhealth2023yang}

\paragraph*{Prompting}
Few-shot prompting~\citep{brown2020language} is used to predict \ac{sdoh} labels from clinical documentation.
Our prompts' role is a certified medical coder, specify labeling criteria, and include coding example (both positive and negative). Our prompts are given in Appendix~\ref{app:final-prompt}.

\paragraph*{Evaluation}
We evaluated seven \acp{llm} spanning frontier closed-source, open-source, reasoning-capable, and traditional architectures on multi-label classification of V-codes that represent adverse \ac{sdoh} labels.  We processed each clinical note through the model independently across admission, then aggregated predictions at the admission level.




\section{Results}

The results of the models we tested are in Table~\ref{tab:overall_models}.
GPT-5-mini achieved the highest overall performance with 72.2\% exact match and 89.1\% micro-F1, which demonstrates frontier models' capability in extracting V-codes in clinical documentation. Open-source models demonstrated competitive capabilities, with Qwen3-32B and GPT-OSS-20B achieving exact match scores of 65.5\% and 65.2\%, respectively. This performance suggests that effective \ac{sdoh} prediction does not require proprietary systems.

Traditional language models showed significantly lower performance, with Llama-3.3-70B and GPT-3.5 achieving exact match scores of 35.5\% and 14.4\%, respectively. The performance difference between traditional and reasoning-capable models indicates that multi-label \ac{sdoh} classification presents challenges that benefit from advanced reasoning capabilities, particularly when processing lengthy clinical narratives with distributed evidence.

Reasoning models leverage the chain-of-thought, which is a series of intermediate reasoning steps the model performs before producing an output~\citep{chainofthought}. These models perform better given the length of the prompt, and complexity of conjunctions (\mbox{i.e.} inclusion and exclusion clauses), and negative examples.  For example, our prompt contains hierarchical code exclusion rules (such as identifying \vname{homelessness} in an example, but instructing the exclusion of \vname{inadequate material resources}).


\begin{table}[b!]
\centering
\tabsize
\begin{tabular}{lcc}
\toprule
\textbf{Metric} & \textbf{Amended} & \textbf{Non-Amended} \\
\midrule
Exact Match & 72.2 & 48.9 \\
Macro-F1    & 66.5 & 32.2 \\
Micro-F1    & 89.1 & 67.0 \\
\bottomrule
\end{tabular}
\caption{GPT-5-mini multi-label classification performance against the amended codes versus non-amended diagnoses codes}
\label{tab:dual_evaluation}
\end{table}
\paragraph*{Label-Specific Performance}
GPT-5-mini's performance varied across \ac{sdoh} categories (Table~\ref{tab:per_code_performance}). The model achieved high performance for \vname{homelessness} and \vname{family alcoholism}, where documentation typically follows consistent terminology. Moderate performance was observed for \vname{unemployment} and \vname{inadequate material resources}. Performance decreased for labels requiring inference across scattered evidence, including \vname{no family caregiver}, \vname{legal problems}, and \vname{emotional abuse}. While this pattern suggests labels with explicit documentation are more reliably predicted, the conservative prompting strategy for codes with low support may have limited the model's ability to capture more nuanced cases, and further prompt optimization could improve performance on complex inference tasks.

\paragraph*{Evaluation Against Original Coding}
We evaluated model performance on the amended codes and the non-amended codes (Table~\ref{tab:dual_evaluation}). Evaluated on the amended codes, GPT-5-mini achieved 72.2\% exact match and 89.1\% micro-F1. Performance against non-amended codes showed 48.9\% exact match and 67.0\% micro-F1. The difference in performance reflects instances where clinical notes contained evidence for additional \ac{sdoh} labels beyond those originally diagnosed.

\section{Discussion}

The evaluation metrics support that \acp{llm} can predict adverse \ac{sdoh} V-codes from clinical notes with high accuracy (89.1\% micro-F1). Furthermore, we identified clinical notes that contained adverse \ac{sdoh} semantics beyond those originally diagnosed.  The performance difference between the prediction of amended codes and the non-amended codes suggests non-agreement in the diagnosis of V-codes among physicians. There is potential for these multi-label classification models to bridge the gap between unstructured documentation of \acp{sdoh} and missed diagnoses. 


The model performed best on labels with standard terminology and struggled with recall when documentation varied, such as \vname{history of emotional abuse}.
However, further experimentation is needed improve precision-recall tradeoff. Open-source models demonstrated competitive performance, suggesting effective V-code prediction may be achievable without requiring commercial models.

\section{Conclusion}
Reasoning \acp{llm} can effectively predict adverse \ac{sdoh} labels from clinical documentation through few-shot prompting approaches. The competitive performance of open-source models compared to commercial models demonstrates that effective \ac{sdoh} prediction does not require proprietary systems, making these capabilities accessible to resource-constrained healthcare settings. Our identification of admissions with documented but uncoded \ac{sdoh} factors highlights a gap between clinical documentation and coding practices.
As healthcare systems increasingly recognize \ac{sdoh} as critical quality indicators, classification models can improve both clinical care and population health management.

\section{Limitations}

The use of MIMIC-III admissions potentially limits generalizability to other institutions with varied documentation practices. The study examined eight ICD-9 V-codes, a subset of available \ac{sdoh} codes. The evaluation dataset was developed by a single annotator without an inter-rater reliability assessment.

The notes used in our experiments are limited to those exported in the MIMIC-III dataset, which include only discharge summaries and those from the ICU department~\cite{landesNewPublicCorpus2022}.  Notes during the hospital stay for patients that are moved to other departments during their recovery are omitted and likely have an impact on model performance.

\section{Acknowledgments}

This work was funded by an award from the Center for Health Equity using Machine Learning and Artificial Intelligence (CHEMA) postdoctoral funding award at the University of Illinois Chicago.

\label{sec:cite}
\bibliography{jmlr-sample}

@article{Torres2017ICDSocialCodes,
  author  = {Torres, Jacqueline M. and Lawlor, John and Colvin, Jeffrey D. and Sills, Marion R. and Bettenhausen, Jessica L. and Davidson, Amber and Cutler, Gretchen J. and Hall, Matt and Gottlieb, Laura M.},
  title   = {ICD Social Codes: An Underutilized Resource for Tracking Social Needs},
  journal = {Medical Care},
  year    = {2017},
  volume  = {55},
  number  = {9},
  pages   = {810--816},
  doi     = {10.1097/MLR.0000000000000764}
}

@article{Hatef2019SBDHAvailability,
  author  = {Hatef, Elham and Rouhizadeh, Masoud and Tia, Iddrisu and Lasser, Elyse and Hill-Briggs, Felicia and Marsteller, Jill and Kharrazi, Hadi},
  title   = {Assessing the Availability of Data on Social and Behavioral Determinants in Structured and Unstructured Electronic Health Records: A Retrospective Analysis of a Multilevel Health Care System},
  journal = {JMIR Medical Informatics},
  year    = {2019},
  volume  = {7},
  number  = {3},
  pages   = {e13802},
  doi     = {10.2196/13802}
}

@article{Chen2020SDOHReview,
  author    = {Chen, Min and Tan, Xuemei and Padman, Rema},
  title     = {Social determinants of health in electronic health records and their impact on analysis and risk prediction: A systematic review},
  journal   = {Journal of the American Medical Informatics Association},
  volume    = {27},
  number    = {11},
  pages     = {1764--1773},
  year      = {2020},
  month     = nov,
  doi       = {10.1093/jamia/ocaa143},
  pmid      = {33202021},
  pmcid     = {PMC7671639}
}

@inproceedings{Mullenbach2018CAML,
  author    = {Mullenbach, James and Wiegreffe, Sarah and Duke, Jon and Sun, Jimeng and Eisenstein, Jacob},
  title     = {Explainable Prediction of Medical Codes from Clinical Text},
  booktitle = {Proceedings of NAACL-HLT 2018},
  year      = {2018},
  pages     = {1101--1111},
  address   = {New Orleans, LA},
  publisher = {Association for Computational Linguistics},
  doi       = {10.18653/v1/N18-1100},
  url       = {https://aclanthology.org/N18-1100/}
}

@article{Johnson2016MIMICIII,
  author  = {Johnson, Alistair E. W. and Pollard, Tom J. and Shen, Lu and Lehman, Li-wei H. and Feng, Mengling and Ghassemi, Mohammad and Moody, Benjamin and Szolovits, Peter and Celi, Leo Anthony and Mark, Roger G.},
  title   = {MIMIC-III, a Freely Accessible Critical Care Database},
  journal = {Scientific Data},
  year    = {2016},
  volume  = {3},
  pages   = {160035},
  doi     = {10.1038/sdata.2016.35}
}

@inproceedings{brown2020language,
  author    = {Tom B. Brown and Benjamin Mann and Nick Ryder and Melanie Subbiah 
               and Jared Kaplan and Prafulla Dhariwal and Arvind Neelakantan 
               and Pranav Shyam and Girish Sastry and Amanda Askell and 
               Sandhini Agarwal and Ariel Herbert-Voss and Gretchen Krueger 
               and Tom Henighan and Rewon Child and Aditya Ramesh and 
               Daniel M. Ziegler and Jeffrey Wu and Clemens Winter and 
               Christopher Hesse and Mark Chen and Eric Sigler and Mateusz Litwin 
               and Scott Gray and Benjamin Chess and Jack Clark and 
               Christopher Berner and Sam McCandlish and Alec Radford and 
               Ilya Sutskever and Dario Amodei},
  title     = {Language Models are Few-Shot Learners},
  booktitle = {Proceedings of the 34th International Conference on 
               Neural Information Processing Systems},
  series    = {NeurIPS '20},
  year      = {2020},
  publisher = {Curran Associates Inc.},
  address   = {Red Hook, NY, USA}
}

@article{chainofthought,
      title={Chain-of-Thought Prompting Elicits Reasoning in Large Language Models}, 
      author={Jason Wei and Xuezhi Wang and Dale Schuurmans and Maarten Bosma and Brian Ichter and Fei Xia and Ed Chi and Quoc Le and Denny Zhou},
      year={2023},
      eprint={2201.11903},
      archivePrefix={arXiv},
      primaryClass={cs.CL},
      url={https://arxiv.org/abs/2201.11903}, 
}

@article{Guevara2024LLMSDoHnpjDM,
  author  = {Guevara, Marco and Chen, Shan and Thomas, Spencer and Chaunzwa, Tafadzwa L. and Franco, Idalid and Kann, Benjamin and Moningi, Shalini and Qian, Jack and Goldstein, Madeleine and Harper, Susan and Aerts, Hugo J. W. L. and Savova, Guergana K. and Mak, Raymond H. and Bitterman, Danielle S.},
  title   = {Large Language Models to Identify Social Determinants of Health in Electronic Health Records},
  journal = {npj Digital Medicine},
  year    = {2024},
  volume  = {7},
  number  = {1},
  pages   = {—},
  doi     = {10.1038/s41746-023-00970-0}
}

@article{Landes2025HybridSDOH,
  author  = {Landes, Paul and Sun, Jimeng and Cross, Adam},
  title   = {Integration of Large Language Models and Traditional Deep Learning for Social Determinants of Health Prediction},
  journal = {arXiv preprint arXiv:2505.04655},
  year    = {2025},
  url     = {https://arxiv.org/abs/2505.04655}
}

@techreport{MIInitiatives2024SDOH,
  title        = {Social Determinants of Health: Discussion Paper 101 for Health Care},
  author       = {{MI Multipayer Initiatives}},
  year         = {2024},
  institution  = {MI Multipayer Initiatives},
  url          = {https://mimultipayerinitiatives.org},
  note         = {Accessed: 2025-08-25}
}

@article{clinicalbert,
  title   = {Publicly Available Clinical {BERT} Embeddings},
  author  = {Alsentzer, Emily and Murphy, John R. and Boag, Willie and Weng, Wei-Hung and Jin, Di and Naumann, Tristan and McDermott, Matthew B. A.},
  journal = {arXiv preprint arXiv:1904.03323},
  year    = {2019},
  url     = {https://arxiv.org/abs/1904.03323}
}

@inproceedings{pyhealth2023yang,
    author = {Yang, Chaoqi and Wu, Zhenbang and Jiang, Patrick and Lin, Zhen and Gao, Junyi and Danek, Benjamin and Sun, Jimeng},
    title = {{PyHealth}: A Deep Learning Toolkit for Healthcare Predictive Modeling},
    url = {https://github.com/sunlabuiuc/PyHealth},
    booktitle = {Proceedings of the 27th ACM SIGKDD International Conference on Knowledge Discovery and Data Mining (KDD) 2023},
    year = {2023}
}

@inproceedings{landesNewPublicCorpus2022,
  title = {A {{New Public Corpus}} for {{Clinical Section Identification}}: {{MedSecId}}},
  shorttitle = {{{MedSecID}}},
  booktitle = {Proceedings of the 29th {{International Conference}} on {{Computational Linguistics}}},
  author = {Landes, Paul and Patel, Kunal and Huang, Sean S. and Webb, Adam and Di Eugenio, Barbara and Caragea, Cornelia},
  date = {2022-10-02},
  pages = {3709--3721},
  publisher = {International Committee on Computational Linguistics},
  location = {Gyeongju, Republic of Korea},
  url = {https://aclanthology.org/2022.coling-1.326},
  urldate = {2022-10-13},
  year={2022},
  eventtitle = {{{COLING}} 2022}
}

\clearpage
\onecolumn
\appendix

\section{Final Prompt for \ac{sdoh} Coding}
\label{app:final-prompt}

\begin{lstlisting}
# SDOH Medical Coding Task

You are an expert medical coder. Your job is to find
Social Determinants of Health (SDOH) codes in clinical notes.

## CRITICAL PREDICTION RULES:
1. IF YOU FIND EVIDENCE FOR A CODE -> YOU MUST PREDICT THAT CODE
2. BE CONSERVATIVE: Only predict codes with STRONG evidence
3. Evidence detection = Automatic code prediction

## CODE HIERARCHY - ABSOLUTELY CRITICAL:
HOMELESS (V600) COMPLETELY EXCLUDES V602:
- [X] If you see homelessness -> ONLY predict V600, NEVER V602
- [X] "Homeless and can't afford X" -> STILL only V600
- [OK] V602 is EXCLUSIVELY for housed people with explicit financial hardship

V602 ULTRA-STRICT REQUIREMENTS:
- [OK] Must have exact quotes: "cannot afford medications", "no money for food"
- [X] NEVER infer from: homelessness, unemployment, substance use, mental health
- [X] NEVER predict from: "poor", "disadvantaged", social circumstances

## Available ICD-9 Codes:

Housing & Resources:
- V600: Homeless (living on streets, shelters, hotels, motels, temporary housing)
- V602: Cannot afford basic needs. ULTRA-STRICT RULES:
  - [OK] ONLY if EXPLICIT financial statements: 
        "cannot afford medications", "unable to pay for treatment", 
        "no money for food", "financial hardship preventing care"
  - [X] NEVER predict from social circumstances: substance abuse, unemployment, 
        mental health, divorce
  - [X] NEVER predict from discharge inventory: "no money/wallet returned"
  - [X] NEVER predict for insurance/benefits mentions: "on disability", "has Medicaid"
- V604: No family/caregiver available. CRITICAL RULES:
  - [X] NOT just "lives alone" or "elderly"
  - [OK] ONLY if: "no family contact" AND "no support" AND "no one to help"
  - [X] NOT for: "lives alone but daughter visits", "independent"

Employment & Legal:
- V620: Unemployed (current employment status) -- EXPLICIT ONLY
  - [OK] Predict only if one of these exact employment-status phrases appears:
    unemploy(ed|ment), jobless, no job, out of work, without work, 
    not working, between jobs, on unemployment, receiving unemployment,
    laid off, fired, terminated, lost job
  - [X] Do NOT infer from "used to work", "formerly employed", etc.
  - [X] Contradictions (block V620 if nearby): "employed", "works at", "working", etc.
  - [X] Non-employment uses of "work": "work of breathing", "social work", etc.
  - [X] Exclusions: retired, student, stay-at-home parent, medical leave
- V625: Legal problems. ACTIVE LEGAL PROCESSES ONLY:
  - [OK] Criminal: arrest, jail, prison, parole, probation, charges, court case
  - [OK] Civil: restraining order, custody case, litigation
  - [OK] Guardianship: legal capacity hearing, court-appointed guardian
  - [OK] Child welfare: CPS/DCF removal, court-ordered custody
  - [X] Do NOT code for history of crime, substance use, or homelessness 
        without legal process

Health History:
- V1541: Physical/sexual abuse (violence BY another person). CRITICAL:
  - [OK] Physical violence, sexual abuse, assault, domestic violence, rape
  - [X] NOT accidents, falls, or when patient is aggressor
- V1542: Pure emotional/psychological abuse. MUTUALLY EXCLUSIVE WITH V1541:
  - [OK] ONLY if NO physical/sexual abuse mentioned AND explicit emotional abuse:
        witnessed violence, verbal abuse, psychological abuse, 
        emotional manipulation, intimidation
  - [X] NEVER predict if physical/sexual abuse mentioned
  - [X] NEVER predict both V1541 AND V1542 for same patient
  - [X] Depression/anxiety/suicidality alone = NO CODE

Family History:
- V6141: Family alcoholism
  - [OK] Kinship + alcohol terms (father alcoholic, mother ETOH abuse, etc.)
  - [X] Negations: "denies family history of alcoholism"
  - [X] NEVER for patient's own history

## ENHANCED NEGATIVE EXAMPLES:
V602 FALSE POSITIVES TO AVOID:
- [X] "Homeless patient" -> V600 ONLY
- [X] "Lives in shelter, gets food stamps" -> V600 ONLY
- [X] "Homeless, on disability" -> V600 ONLY
- [X] "No permanent address, has Medicaid" -> V600 ONLY
- [X] "Homeless and can't afford medications" -> V600 ONLY
- [X] "Lives in poverty" -> NO CODE
- [X] "Financial strain from divorce" -> NO CODE

V604 FALSE POSITIVES TO AVOID:
- [X] "82 year old lives alone" -> NO CODE
- [X] "Lives by herself" -> NO CODE
- [X] "Widowed, lives alone, son calls daily" -> NO CODE

V1542 FALSE POSITIVES TO AVOID:
- [X] "History of physical and sexual abuse" -> V1541 ONLY
- [X] "PTSD from rape at age 7" -> V1541 ONLY
- [X] "Childhood sexual abuse by uncle" -> V1541 ONLY
- [X] "History of domestic abuse" -> V1541 ONLY
- [X] "Depression and anxiety" -> NO CODE
- [X] "Suicide attempts" -> NO CODE
- [X] "History of abuse" (unspecified) -> NO CODE
- [X] "Recent argument with partner" -> NO CODE

V1542 TRUE POSITIVES TO CAPTURE:
- [OK] "Witnessed violence as a child" -> V1542
- [OK] "Emotionally abusive relationship for 14 years" -> V1542
- [OK] "Verbal abuse from controlling partner" -> V1542
- [OK] "Jealous and controlling behavior" -> V1542

## CONFIDENCE RULES:
HIGH CONFIDENCE (Predict):
- Direct statement
- Multiple evidence pieces
- Explicit matching phrase

LOW CONFIDENCE (Don't Predict):
- Ambiguous language
- Single weak indicator
- Contradictory evidence

## Key Rules:
1. Precision over Recall: Better to miss a code than falsely predict
2. Evidence-Driven: Strong evidence required for prediction
3. Multiple codes allowed: But each needs independent evidence
4. Conservative approach: When in doubt, don't predict

## Output Format:
Return applicable codes separated by commas, or "None".

Examples:
V600, V625
None
\end{lstlisting}
\begin{figure}[H] 
\centering
\caption{Final optimized prompt used for admission-level \ac{sdoh} ICD-9 V-code prediction. 
The prompt encodes hierarchy rules, precision-first criteria, negative examples, and explicit output formatting. 
This full text is provided to ensure reproducibility.}
\end{figure}
\section{Annotation Guide for \ac{sdoh} ICD-9 V-codes}
\label{app:annotation-guide}
\subsection{V60.0: Homelessness}
\textbf{Definition:} Patient has no permanent residence or lives in transient arrangements such as streets, shelters, or motels. \\
\textbf{Approximate Synonyms:} Homeless, currently homeless. \\
\textbf{Applies To:} Hobos, tramps, vagabonds, social migrants, transients. \\
\textbf{Inclusion Criteria:} Explicit statements such as ``Patient is homeless,'' ``Lives in shelter,'' or ``No permanent address.'' \\
\textbf{Exclusion Criteria:} Financial hardship only (V60.2), statements like ``lives alone'' or ``poor'' without loss of housing. \\
\textbf{Examples:}
\begin{itemize}
\item Positive: ``Lives in a homeless shelter after eviction.''
\item Negative: ``Struggling to pay rent but living with family.''
\end{itemize}

\subsection{V60.2: Inadequate Material Resources}
\textbf{Definition:} Patient lacks financial means for necessities such as food, medication, or housing. \\
\textbf{Approximate Synonyms:} Financial problem. \\
\textbf{Applies To:} Economic problem, poverty not otherwise stated. \\
\textbf{Inclusion Criteria:} Explicit financial statements (e.g., ``Cannot afford medications,'' ``No money for food''). \\
\textbf{Exclusion Criteria:} Any homelessness (V60.0), implicit hardship without direct evidence. \\
\textbf{Examples:}
\begin{itemize}
\item Positive: ``Financial hardship preventing purchase of insulin.''
\item Negative: ``Homeless and can’t afford meds.'' (V60.0 only)
\end{itemize}

\subsection{V60.4: No other household member able to render care}
\textbf{Definition:} Indicates no family or caregiver available to provide care or support. \\
\textbf{Applies To:} Family member too ill or unsuited to render care, partner temporarily away, or person away from usual home. \\
\textbf{Inclusion Criteria:} Explicit lack of family support (e.g., ``No family contact and no one to help with care.''). \\
\textbf{Exclusion Criteria:} ``Lives alone,'' ``independent,'' or ``widowed'' without need for care. \\
\textbf{Examples:}
\begin{itemize}
\item Positive: ``No family or friends available to assist after surgery.''
\item Negative: ``Lives alone, independent in activities of daily life.''
\end{itemize}

\subsection{V62.0: Unemployment}
\textbf{Definition:} Patient currently lacks employment. \\
\textbf{Excludes:} Main problem due to economic inadequacy (V60.2). \\
\textbf{Inclusion Criteria:} Direct mention of unemployment or job loss (``unemployed,'' ``laid off,'' ``no job''). \\
\textbf{Exclusion Criteria:} Retired, student, stay-at-home parent, or on medical leave; ``used to work.'' \\
\textbf{Examples:}
\begin{itemize}
\item Positive: ``Unemployed, seeking work.''
\item Negative: ``Previously worked as nurse.''
\end{itemize}

\subsection{V62.5: Legal Problems}
\textbf{Definition:} Current involvement with legal or judicial processes (criminal, civil, custody, or guardianship). \\
\textbf{Approximate Synonyms:} Arrested, charged with crime, imprisonment, litigation, parole, bail, probation, court case pending. \\
\textbf{Applies To:} Imprisonment, legal investigation, litigation, prosecution. \\
\textbf{Inclusion Criteria:} Active legal process (arrest, trial, custody case, guardianship hearing). \\
\textbf{Exclusion Criteria:} History of crime or incarceration without current legal process. \\
\textbf{Examples:}
\begin{itemize}
\item Positive: ``Currently on probation after DUI arrest.''
\item Negative: ``History of incarceration in 2010.''
\end{itemize}

\subsection{V15.41: History of Physical or Sexual Abuse}
\textbf{Definition:} Past exposure to physical or sexual violence inflicted by another person. \\
\textbf{Approximate Synonyms:} History of physical abuse, sexual abuse, child abuse, domestic violence, rape. \\
\textbf{Applies To:} Rape. \\
\textbf{Inclusion Criteria:} Mentions such as ``History of sexual abuse'' or ``Physically abused by spouse.'' \\
\textbf{Exclusion Criteria:} Accidents, self-injury, patient as aggressor, emotional abuse only (V15.42). \\
\textbf{Examples:}
\begin{itemize}
\item Positive: ``Reports childhood sexual abuse by uncle.''
\item Negative: ``Verbal abuse from partner.'' (V15.42)
\end{itemize}

\subsection{V15.42: History of Emotional Abuse or Neglect}
\textbf{Definition:} History of psychological, emotional, or neglectful abuse without physical or sexual component. \\
\textbf{Approximate Synonyms:} Emotional abuse, neglect, psychological abuse, verbal abuse, child neglect. \\
\textbf{Applies To:} Neglect. \\
\textbf{Inclusion Criteria:} ``Emotionally abusive relationship,'' ``childhood neglect,'' ``verbal abuse.'' \\
\textbf{Exclusion Criteria:} Physical or sexual abuse (V15.41), depression/anxiety/PTSD alone. \\
\textbf{Examples:}
\begin{itemize}
\item Positive: ``Endured years of emotional abuse from spouse.''
\item Negative: ``PTSD from physical assault.'' (V15.41)
\end{itemize}

\subsection{V61.41: Family History of Alcoholism}
\textbf{Definition:} Indicates alcoholism or alcohol dependence in the patient’s family. \\
\textbf{Approximate Synonyms:} Family history of alcohol dependence, FHX of alcoholism, adult child of alcoholic. \\
\textbf{Inclusion Criteria:} Mentions such as ``Father alcoholic,'' ``Mother with ETOH abuse,'' or ``Family history of alcoholism.'' \\
\textbf{Exclusion Criteria:} Patient’s own alcohol use; negations (``Denies family history of alcoholism''). \\
\textbf{Examples:}
\begin{itemize}
\item Positive: ``Mother and uncle both heavy drinkers.''
\item Negative: ``Patient with long history of alcohol use disorder.''
\end{itemize}

\subsection*{General Annotation Rules}
\begin{enumerate}
\item Precision over recall: only label with explicit evidence.
\item Mutual exclusions: V60.0 supersedes V60.2; V15.41 and V15.42 are mutually exclusive.
\item Each code requires independent evidence.
\item When in doubt, do not assign a code.
\end{enumerate}

\begin{figure}[H] 
\centering
\caption{Annotation guide used to label \ac{sdoh} circumstances in clinical text for ICD-9 V-code prediction. 
Each section defines the code, domain, inclusion and exclusion criteria, and examples. 
Annotators must use precision-first judgment and follow hierarchy rules as outlined.}
\end{figure}
\end{document}